\documentclass[sigconf]{acmart}
\usepackage[disable]{todonotes}
\usepackage{hyperref}
\usepackage{caption}
\usepackage{subcaption}
\usepackage{multirow}
\usepackage{lipsum}
\usepackage{threeparttable}
\AtBeginDocument{%
  \providecommand\BibTeX{{%
    \normalfont B\kern-0.5em{\scshape i\kern-0.25em b}\kern-0.8em\TeX}}}

\setcopyright{acmcopyright}
\copyrightyear{2023}
\acmYear{2023}
\acmDOI{XXXXXXX.XXXXXXX}

\acmConference[HIP '23]{7th International Workshop on Historical Document Imaging and Processing (HIP’23)}{August 25,
  2023}{San José, California}
\acmPrice{15.00}
\acmISBN{978-1-4503-XXXX-X/18/06}

\begin{document}

\title{Handwritten Text Recognition from Crowdsourced Annotations}


%\author{Anonymous}
%\affiliation{
%    \country{}
%}

\author{Solène Tarride}
\affiliation{%
  \institution{TEKLIA}
  \city{Paris}
  \country{France}
}
\email{starride@teklia.com}

\author{Tristan Faine}
\affiliation{%
  \institution{TEKLIA, Nantes Université}
  \city{Nantes}
  \country{France}
}

\author{Mélodie Boillet}
\affiliation{%
  \institution{TEKLIA}
  \city{Paris}
  \country{France}
}

\author{Harold Mouchère}
\affiliation{%
  \institution{Nantes Université}
  \city{Nantes}
  \country{France}
}

\author{Christopher Kermorvant}
\affiliation{%
  \institution{TEKLIA}
  \city{Paris}
  \country{France}
}

%\renewcommand{\shortauthors}{xxx et al.}

\begin{abstract}
%Crowdsourcing is a popular way of obtaining ground truth annotations. To ensure the quality of the collected transcriptions, a simple method is to have multiple annotators transcribe the same document and measure their level of agreement.
% experiments
In this paper, we explore different ways of training a model for handwritten text recognition when multiple imperfect or noisy transcriptions are available. 
We consider various training configurations, such as selecting a single transcription, retaining all transcriptions, or computing an aggregated transcription from all available annotations. 
In addition, we evaluate the impact of quality-based data selection, where samples with low agreement are removed from the training set. Our experiments are carried out on municipal registers of the city of Belfort (France) written between 1790 and 1946.
% results
The results show that computing a consensus transcription or training on multiple transcriptions are good alternatives. However, selecting training samples based on the degree of agreement between annotators introduces a bias in the training data and does not improve the results. Our dataset is publicly available on Zenodo.
\end{abstract}




\begin{teaserfigure}
  \includegraphics[width=\linewidth]{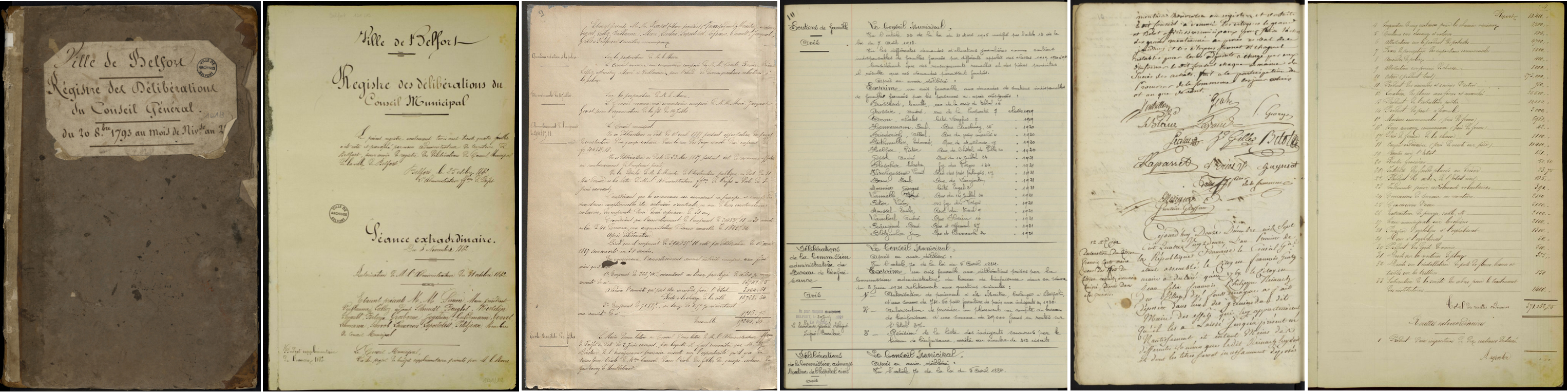}
    \caption{Samples from the minutes of Belfort municipal council drawn up between 1790 and 1946. These documents include deliberations, lists of councillors, convocations, and agendas.}
    \label{fig:samples_belfort}
\end{teaserfigure}

\received{28 April 2023}

\maketitle

\section{Introduction}
The Municipal Archives of Belfort have launched a pilot project for the automatic transcription of all council minutes. The aim of the project is to automatically process 18,500 pages of various documents written between 1790 and 1946, as presented in Figure \ref{fig:samples_belfort}.
Images and transcriptions will then be made freely available so that citizens can not only browse the pages but also search for specific information. This will give citizens the opportunity to participate in local life by discovering how the City Council works and how the city has developed over the centuries.

% Annotation campaign
%To collect training data for automatic Handwritten Text Recognition (HTR) models, we set up an open collaborative annotation campaign. We randomly selected 616 images and used Doc-UFCN \cite{boillet2020} to automatically detect text lines in the pages. Collaborative transcription was carried out using the Callico\footnote{\url{https://callico.teklia.com}} annotation tool, which allows any user to request tasks, and automatically assigns three pages for annotation. The annotation interface is presented in Figure \ref{fig:callico}. To annotate a page, the user must enter the transcription of the line highlighted in blue in the corresponding text box, also highlighted in blue. All lines highlighted in green must be transcribed by the user to complete the task. 

\begin{figure}[ht]
    \centering
    \includegraphics[width=\linewidth]{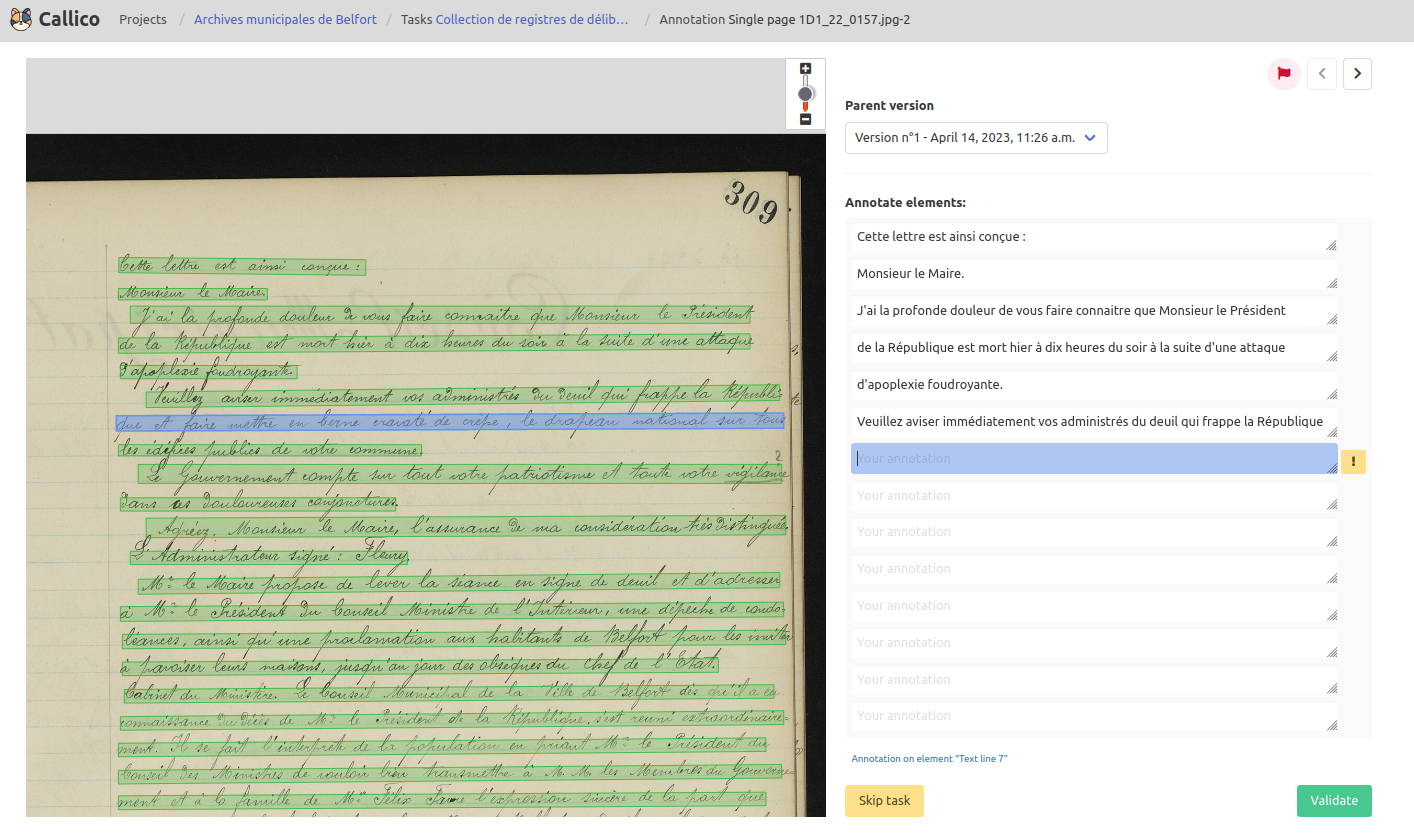}
    \caption{Screenshot of Callico, the annotation tool used for collaborative annotation.}
    \label{fig:callico}
\end{figure}

To collect training data for automatic Handwritten Text Recognition (HTR) models, we set up an open collaborative annotation campaign using the Callico\footnote{\url{https://callico.teklia.com}} annotation tool, which allows any user to request tasks. The annotation interface is presented in Figure \ref{fig:callico}. To annotate a page, the user must enter the transcription of the line highlighted in blue in the corresponding text box, also highlighted in blue. All lines highlighted in green must be transcribed by the user to complete the task. 

% Quality control
Since the model will learn from the collaborative transcriptions, their quality is crucial as it will affect its performance. An easy way to estimate the quality of the collected data is to have two annotators transcribe each line. For this reason, the transcription campaign is configured so that each document can be annotated by two different annotators. As there is no clear way to identify potential transcription issues for pages transcribed by a single annotator, we also rely on pre-trained automatic HTR models to generate two automatic transcriptions for each line. Ultimately, each line will have either one or two human transcriptions and two automatic transcriptions.

% Questions
This paper explores various approaches to train a model when multiple uncertain transcriptions are available for each line. We investigate three main issues: transcription selection, sample selection based on transcription quality, and data splitting. This paper aims to answer the following questions.

\begin{enumerate}
    \item \textbf{How to train an HTR model when multiple annotations are available for each image?} Three main training strategies are explored: merging multiple transcriptions, selecting only one transcription, or using all available transcriptions.  
    \item \textbf{How to identify low-quality annotations, and what is their impact on model performance?} We measure the degree of agreement between multiple transcriptions. Then, we use this metric to discard unreliable samples and measure the impact on model performance.
    \item \textbf{How to partition the data into training, validation and test sets when transcription quality is unknown?} We address the issue of data partitioning in scenarios where the transcriptions are uncertain. Two distinct splits are defined: the first ensures the reliability of the validation and test sets, while the second is randomized.
\end{enumerate}
  %  \item is it better to train a model using a smaller number of reliable samples or a larger number of uncertain samples?

%\todo[inline]{HM: dire tout de suite les Input/output attendus: il y a 4 sources à fusionner: 2 annotateurs et 2 IA, et en output une proposition de fusion et surtout un score "agreement".}

This paper is organized as follows. The next Section \ref{sec:related_works} presents related works. Section \ref{sec:dataset} introduces the dataset and Section \ref{sec:experiments} presents our experiments and results. Finally, we discuss the results and conclude in Section \ref{sec:conclusion}.

\section{Related Works} 
\label{sec:related_works}

In this paper, we focus on two main points. First, we explore the different methods for quality assessment in the context of crowdsourcing. Second, we investigate potential training strategies when multiple noisy labels are available for each sample. The literature on each of these points is detailed in the following paragraphs.

\subsection{Label quality assessment}

Crowdsourcing platforms are a relatively new way for task providers, platform managers and workers to interact together in order to meet the task provider's expectations. 
The survey proposed by Daniel \textit{et al.} \cite{Daniel_2018} provides comprehensive insights into quality assessment in the context of crowdsourcing.
%there are no enforced robust and flexible quality control framework for these platforms, as current research is more focused on individual aspects like worker reputation, or task redundancy.
% Still, quality assessment of multiple answers is important% in order to be able to define a consensus.

\subsubsection{Group assessment}
The simplest method of ensuring high quality annotations is to allow humans to agree on the best label \cite{Daniel_2018} by allowing voting strategies, peer review mechanisms, or output agreement. Collaborative annotation has been shown to be more efficient than aggregating individual annotations, as demonstrated by Blickhan \textit{et al.} \cite{zoouniverse}, which suggests that annotators should be allowed to see previous annotations made by others. 
Another approach consists in relying on moderation to correct or discard unreliable transcriptions. 

Human-based approaches to quality control are extremely effective, but require considerable time and effort to implement. In addition, additional features need to be incorporated into the annotation interface to facilitate human-based quality control.

\subsubsection{Computational assessment}

Computational methods can also be used for quality assessment \cite{Daniel_2018}.
A simple method is to annotate a random subset of the dataset with gold truth annotations and compute metrics for quality control. Other approaches rely on content analysis to estimate task difficulty and worker reliability.
Zhu \textit{et al.} \cite{Zhu2021} propose an outlier detection method for classification tasks to detect where annotators disagree. Based on the assumption that similar samples are more likely to share the same label, their method compares each noisy label with the labels of its nearest neighbours in the feature space.
% voting technique based on the distance between a label and the distance to the samples within the cluster for the same label could lead to the detection of corrupted patterns.

%data-centric method to detect 'corrupted patterns', cases where annotators are not in agreement, simply by using the annotations' content and the information contained within the instances, no model required.

\subsection{Learning with multiple labels}
%Taking inspiration from the text summarization task, a consensus between texts can be achieved by creating a representation of the content that different annotators agree on, or selecting the text which is most similar to other texts. Such methods are described as extractive or abstractive in the current literature. \cite{ElKassas2021}
There are several methods for training a model when multiple transcriptions are available. It is possible to select a single label using text aggregation methods, or to adapt training strategies to allow multiple labels for each sample.

\subsubsection{Label selection}
%This can be achieved by generating a representation of the content that most annotators agree on, or by selecting the text that is most similar to other texts.

The task of text summarization can provide some insight into how consensus between different texts can be calculated, using abstractive or extractive methods \cite{ElKassas2021}. 

\textit{Abstractive aggregation}  techniques produce a transcription on which most annotators agree. The consensus text combines the features of each transcription and can be seen as a weighted average of the different transcriptions. Recognizer Output Voting Error Reduction (ROVER) \cite{Fiscus1997}, is a system that uses dynamic programming to align word sequences then applies a voting strategy on every edge of the resulting Word Transition Network. Although ROVER was originally designed to align words, it can also be used at character or sub-word level. The ROVER algorithm is included in the \texttt{crowd-kit} Python package \cite{CrowdKit}. 
Collatex \cite{Collatex} is another tool designed for collating textual sources, it provides different alignment algorithms in order to align tokens, then a voting technique can be applied to the resulting alignment table.

\textit{Extractive aggregation}  techniques select the transcription that is most similar to each other. While this doesn't really produce a consensus, it could be used to discard unlikely annotations. The Reliability Aware Sequence Aggregation (RASA) \cite{Li2019} and the Hybrid Reliability and Representation Aware Sequence Aware Aggregation (HRRASA) \cite{Li2020} algorithms have also been proposed to determine which of the annotators' answers is closest to the estimated aggregated answer. Both methods are able to take into account the reliability of each annotator, which is a representation of how close they have been to previous aggregations over time.
HRRASA differs from RASA in that it considers local reliability within predefined subsets of transcriptions. Both algorithms are implemented in the \texttt{crowd-kit} Python package \cite{CrowdKit}.

\subsubsection{Training strategies that are able to deal with multiple noisy labels}

%In our study, each line is transcribed by at most two different human annotators and two different HTR models. All these transcriptions potentially include noise (typing error, reading difficulty, standardization issues).

%Different profiles and skill levels means that some outputs are partly a result of bias instead of purely being inferred from the data.

While the literature on crowdsourced text transcription is sparse, there are many articles focusing on crowdsourced labels for classification tasks.

%\subsubsection{Deep learning with multiple labels}
%In the case of crowdsourcing, each annotation comes from a different profile, learning the ways in which they differ and recognizing the cases where they are similar can be helpful information that acts as a substitute for gold labels.
\textit{Training with all available labels} is a straightforward and workable solution, as shown by Wei \textit{et al.} \cite{Wei2022b}. The authors assume that instances can be associated with different noisy labels due to different interpretations. According to their results on several datasets, the use of multiple individual labels leads to an increase in accuracy when the number of annotators is insufficient or when the noise rate is high (low-quality annotations).
%Removing annotator identification means that only one noise transition matrix is needed, although a custom per-sample loss function is required. 

\textit{Adapting models to handle multiple noisy labels} is also possible. 
%Some techniques are designed to characterize the noise structure causing annotators to be wrong sometimes, while other techniques try to achieve noise robustness without explicitly labelling it.
CrowdLayer \cite{rodrigues2017deep} relies on a crowd layer that takes the softmax outputs of a standard deep neural network as the input, and learns multiple parametric annotator-specific transition matrices to fit the given label of each annotator independently. UnionNet \cite{Wei2022} is a deep neural network that can be trained directly by maximizing the likelihood of a combined set of labels using a parametric transition matrix, thus lowering the computational cost and possibly learn inter-annotator relationships.
In addition to annotator-based noise, researchers also highlight the impact of sample-based noise \cite{Gao2022, Wei2021}. Gao \textit{et al.} \cite{Gao2022} propose to add regularization terms in the loss function to make weight vectors and confusion matrices sample-dependent.

\subsection{Discussion}
The literature review identifies two main issues that are relevant to our application.

\subsubsection{Quality assessment}
%Human-based quality assessment cannot be done in our case as the interface does not include required features.
As only two human annotations are available for each line, we also use two pre-trained models to generate automatic transcriptions. As a result, four annotations are available for each sample. This makes it possible to compute an agreement score between individual annotations, which can help to estimate the task difficulty and label noise.
%There were no plans for a human-based quality assessment for this campaign, as this would have been too time-consuming. 
%However, as multiple annotations are available for each sample, it is possible to compute an agreement score between individual annotations. This can help estimate the task difficulty and label noise.

\subsubsection{Learning from multiple transcriptions}
Existing methods are designed for classification tasks and cannot be directly applied to handwritten text recognition. In addition, many of these methods require the identification of the annotator, which is not currently possible with our annotation campaign.
However, label selection techniques are interesting because they can potentially correct local errors made by individual annotators. 

\section{Collaborative annotation of the Belfort dataset}
\label{sec:dataset}

In this section, we describe how the transcriptions were collected and present the annotated Belfort dataset. We first randomly selected 616 images and used a pre-trained Doc-UFCN \cite{boillet2020} model to automatically detect text lines in the pages. We then created a collaborative annotation campaign to collect training data for the text recognition models.

\subsection{Annotation guidelines}

The following annotation guidelines have been established in order to make the collected transcriptions as consistent as possible. 

Only highlighted lines should be transcribed. If there is a line segmentation error where the highlighting covers several lines, the transcription should be left blank. If several words are illegible, the transcription should be left blank. Illegible signatures should not be annotated. 
Spelling and conjugation errors should be corrected, punctuation and capitalization should be restored, and abbreviations should be expanded. 
If there is any uncertainty about the transcription, it should be entered as accurately as possible, without adding any characters, and the line should be marked as "uncertain". If the quality of the page is poor and illegible, it should be skipped.

%\begin{itemize}
%    \item All lines highlighted must be transcribed. Areas of the page that are not highlighted should not be transcribed.
%    \item After selecting a line, the text highlighted in blue should be entered in the corresponding annotation box.
%    \item If the highlighting covers multiple lines (line segmentation error), the transcription should be left blank.
%    \item Spelling and conjugation errors should be corrected.
%    \item Punctuation and capitalization should be restored.
%    \item Abbreviations should be expanded.
%    \item In case of doubt, the transcription should be entered as accurately as possible, without adding characters. The line should then be marked as "uncertain" by clicking on the button.
%    \item If several words are illegible, the transcription should be left empty.
%    \item Illegible signatures should not be annotated.
%    \item In case of poor quality and unreadable pages, the page should be skipped.
%    \item Major issues should be reported with the red flag.
%\end{itemize}

At the time we extracted the data from the annotation platform, 64\% of the pages were transcribed by two independent annotators, 27\% by a single annotator, and 9\% were skipped. On average, an annotator took 21 minutes to annotate a full page.

\subsection{Description of the annotated dataset}
From these annotations, we build a line-level dataset for handwritten text recognition. This dataset is publicly available on Zenodo\footnote{\url{https://zenodo.org/record/8041668}}.

\subsubsection{Dataset}
24,105 text lines were transcribed, including 37\% transcribed by two different human annotators. 
A significant gap can be observed between the number of pages with two annotations (64\%) and lines with two annotations (37\%). 
This is because many lines were difficult to read, and were therefore skipped, left blank or ignored by at least one annotator. 
Lines with no transcription are excluded from the dataset. 

\subsubsection{Automatic transcriptions}
We want to investigate training strategies when multiple annotations are available for the same image. However, some lines in our dataset were transcribed by a single annotator. For this reason, we also use two handwriting recognition models (PyLaia \cite{PyLaia} and DAN \cite{DAN}) trained on an earlier version of this dataset to generate automatic annotations. 

\begin{figure}[th!]
     \centering
     \includegraphics[width=\linewidth]{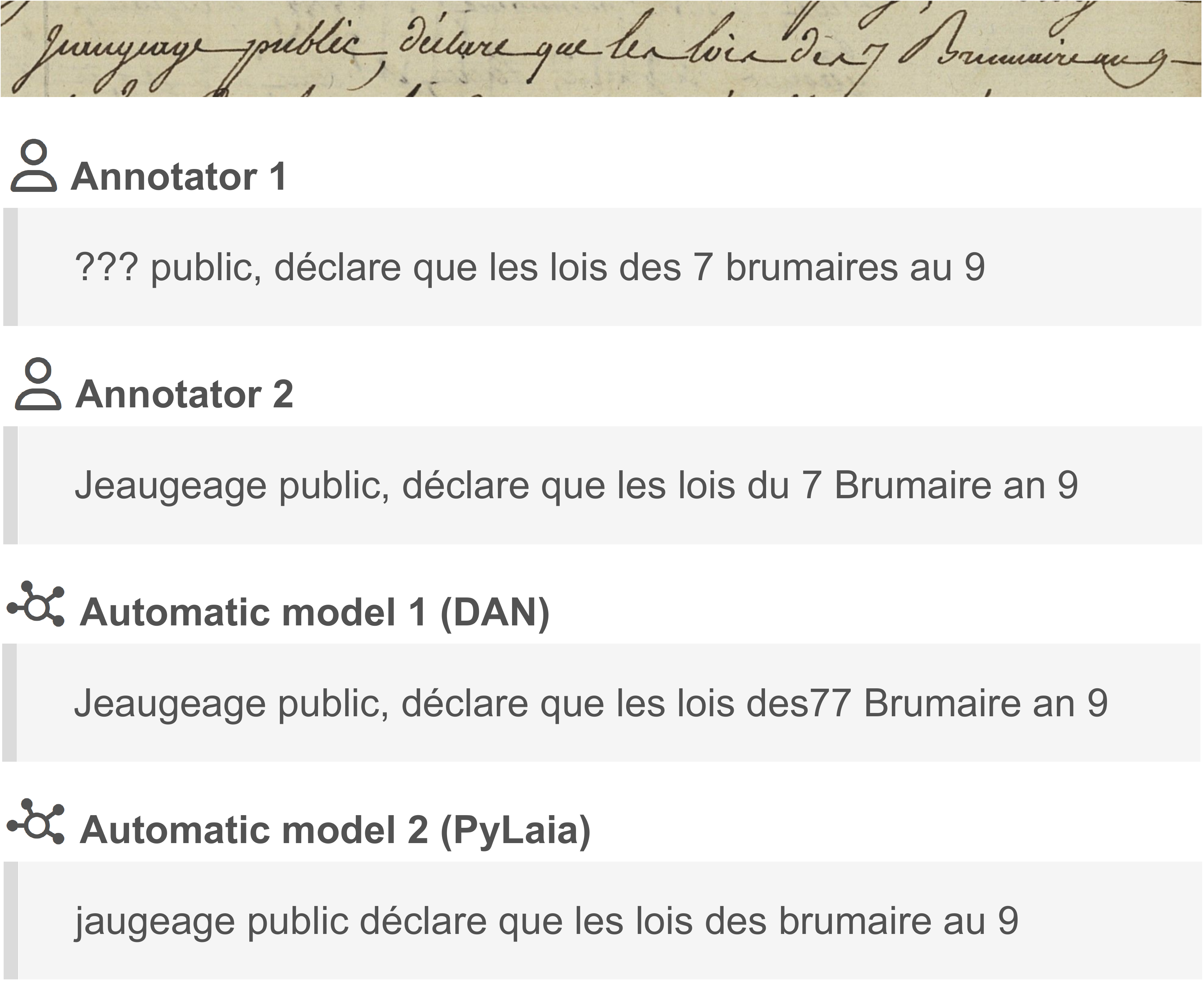}
    \caption{Different transcriptions associated to a single text line. None of these transcriptions are exactly correct. The gold transcription should be "\textit{Jeaugeage public, déclare que les lois des 7 Brumaire au 9}".}
    \label{fig:transcriptions}
% https://demo.arkindex.org/element/669e7b54-b1a6-4ee8-bc46-12297536cf39?from=8386385d-d200-4df4-a7eb-6f6a21bae9cc&highlight=038c9e0c-819a-47b6-84c8-7771e43061c1
\end{figure}

As a result, each line now have three or four transcriptions. These transcriptions often include small errors, as illustrated in Figure \ref{fig:transcriptions}. 

\subsubsection{Splitting strategy}
\label{sec:splitting_strategy}
Since our goal is to train and compare different models, we need a reliable test set to evaluate the models as accurately as possible. However, we have no prior information about the quality of the transcriptions, so there is no guarantee that the annotations are correct. When two manual annotations are available, one way to ensure the quality of the collected transcriptions is to look at the agreement level between the two annotators.

Therefore, the test set was designed to contain only lines where the two annotators agree exactly (Character Error Rate = 0\%). As the validation set is also important when training machine learning models, lines with good agreement (0\% < Character Error Rate < 5\%) between the two human annotations are selected to be in the validation set. Finally, all the other lines are used for training. 

The detailed split is described in Table \ref{tab:split}. We discuss the potential bias introduced by this splitting strategy in Section \ref{sec:bias}.

\begin{table}[htb]
    \centering
    \caption{Details of the data splitting at line level}
    \begin{tabular}{l|rr}
    \toprule
        \bf Split & \bf Number & \bf Percentage (\%)  \\
    \midrule
        Train & 19,013 & 78.9 \\
        Validation & 2,262 & 9.4 \\
        Test & 2,830 & 11.7 \\
    \midrule
        Total & 24,105 & 100.0\\
    \bottomrule
    \end{tabular}
    \label{tab:split}
\end{table}

\subsubsection{Level of agreement between annotators}
The agreement score between multiple transcriptions can be computed using the following approach. First, a consensus transcription is obtained by taking a majority vote at each character position among the N transcriptions, using the ROVER algorithm from the \texttt{crowd-kit} Python package\footnote{\url{https://toloka.ai/docs/crowd-kit/reference/crowdkit.aggregation.texts.rover.ROVER/}} at the character-level. Next, the character distance between each transcription and the ROVER consensus is calculated, to measure the dissimilarity between them. Finally, the computed character distances are averaged across all transcriptions to obtain an agreement score which reflects the overall level of agreement between the transcriptions.

%\begin{itemize}    
%    \item Show examples of high/low agreement. 
%    \item Compare distance to consensus for DAN, PyLaia, Human annotators. 
%\end{itemize}

The main factors contributing to low agreement include poor-quality images, line segmentation errors (mostly merged lines), challenging handwriting, and non-compliance with annotation guidelines (incomplete transcription, punctuation errors, casing errors, additional symbols (such as \textit{"?"}) for uncertain words).  

\section{Experiments}
\label{sec:experiments}

Our experiments focus on three research questions. 
First, we explore different strategies for training a handwriting recognition model when multiple transcriptions are available for each image.
We also measure the impact of data quality on model performance by filtering out unreliable samples. 
Finally, we consider the potential bias introduced by our data splitting strategy. 

\subsection{Experimental setup}

For all the following experiments, we train a PyLaia\footnote{\url{https://github.com/jpuigcerver/PyLaia}} \cite{PyLaia} model consisting of four convolutional blocks and three recurrent blocks, followed by a linear layer. 
Each convolution block is composed of a convolution layer, batch normalization, the LeakyReLU activation function and max pooling. Each recurrent block includes a Bi-LSTM layer.
All images are resized to a fixed size of 128 pixels, while preserving the original aspect ratio. Training is performed using a batch size of 8. 
Early stopping is applied after 50 epochs with no improvement on the validation set.

The model is evaluated on the test set using human transcriptions. 

\subsection{Learning from multiple transcriptions}

We want to compare different ways of integrating multiple transcriptions into the training of a model. We investigate and compare three methods: selecting a single annotation for each line or keeping multiple annotations for each line.

\subsubsection{Selection of a single annotation}
The first approach consists in selecting a single annotation for each line. We compare three ways of doing this: 
\begin{itemize}
    \item Randomly selecting a transcription;
    \item Selecting the best transcription using the RASA algorithm; %(ex Figure 3:     \textit{"Jeaugeage public, déclare que les lois des77 Brumaire an 9"})
    \item Selecting the aggregated transcription obtained using the ROVER algorithm. % (ex Figure 3: \textit{"Jeaugeage public, déclare que les lois des 7 brumaire au 9"})
\end{itemize}

\subsubsection{Retention of multiple annotations} 
The second approach consists in duplicating images with multiple transcriptions, so that a single line appears $n$ times in the training set if it has $n$ different transcriptions. We investigate the following three strategies:

\begin{itemize}
    \item Use all human transcriptions;
    \item Use all human and automatic transcriptions; 
    \item Use all human, automatic, and aggregated transcriptions (RASA and ROVER).
\end{itemize}

\subsubsection{Results} 

\begin{table*}[thb]
    \centering
    \caption{Evaluation results for experiments on transcription selection. The same PyLaia architecture is trained using different transcriptions}
    \label{tab:results_multiple_transcriptions}
    \begin{tabular}{l|cccc}
    \toprule
        \multirow{2}{*}{\bf Transcription selection strategy} & \multicolumn{2}{c}{\bf Validation\footnotemark[1]} & \multicolumn{2}{c}{\bf Test}\\
        & CER (\%) & WER (\%)  & CER (\%) & WER (\%) \\
    \midrule
        Random selection  & 6.04 & 21.53 & 5.44 & 18.37\\
        RASA selection & 4.80 & 17.15 & 4.95 & 17.17 \\
        ROVER consensus & \textbf{4.55} & \textbf{16.73} & 4.95 & 17.08 \\
        All human transcriptions & 6.05  & 22.18 & 5.57 & 19.12 \\
        All human and automatic transcriptions & 6.04 & 20.29 & 5.53 & 16.50 \\
        All transcriptions (human, automatic, RASA, ROVER) & 5.10 & 19.03 & \textbf{4.34} & \textbf{15.14}  \\
    \bottomrule
    \multicolumn{5}{l}{\footnotemark[1] Validation sets include different transcriptions} \\
    \end{tabular}
\end{table*}
The results of our experiments are presented in Table \ref{tab:results_multiple_transcriptions}.

For label selection strategies, RASA and ROVER are better than a random selection on the test set, with the WER reduced by 1 point. Although there is no clear winner between RASA and ROVER, ROVER achieves the best WER. 

For label retention strategies, we observe that including automatic transcriptions helps to reduce the WER. Using all available transcriptions has an even more significant effect, resulting in a decrease of 1 percentage point on the CER and almost 4 points on the WER.

The best overall performance on the test set is achieved by using all available transcriptions, resulting in a 4.34\% CER and a 15.14\% WER. One possible reason for this improvement is that training with a larger number of words in the training set improves PyLaia's implicit language modelling mechanism. 
Despite the fact that each image appears with different labels five or six times in the training set, we observe that the model does not overfit earlier than other training configurations.

\subsection{Quality-driven training sample selection}
In this section, we investigate how transcription quality affects model performance. To achieve this, we use the agreement score to remove unreliable samples from the training set.

\subsubsection{Selection strategy}
Using the ROVER consensus transcription seems to be the best option for training a model. However, if the available transcriptions are very different, the ROVER consensus transcription is likely to contain errors. This can lead to the model being trained with unreliable labels. 
To avoid this, we rely on the agreement score to filter out unreliable training samples (e.g. with low agreement scores) during PyLaia's training, allowing us to evaluate the resulting impact of on performance.

In this study, we experiment with three different thresholds: 90\%, 97\%, and 99\% of agreement score. These thresholds are chosen because they correspond to retaining approximately 75\%, 50\%, and 30\% of the training samples respectively.

\subsubsection{Results}
Results of our experiments are presented in Table \ref{tab:results_sample_selection}.

\begin{table*}[thb]
    \centering
    \caption{Evaluation results for experiments on agreement-based quality selection. The same PyLaia architecture is trained using the ROVER consensus and by filtering out samples with agreement below a certain threshold}
    \label{tab:results_sample_selection}
    \begin{tabular}{l|rrcccc}
    \toprule
        \multirow{2}{*}{\bf Filtering strategy} & \multirow{2}{*}{\bf Threshold} & \multirow{2}{*}{\bf Training samples} & \multicolumn{2}{c}{\bf Validation} & \multicolumn{2}{c}{\bf Test} \\
        & & & CER (\%) & WER (\%) & CER (\%) & WER (\%) \\
    \midrule
        Retain all training samples & 0\% & 19,013 (100.0\%) & \textbf{4.55} & \textbf{16.73} & \textbf{4.95} & \textbf{17.08} \\
        Discard samples with low agreement & 90\% & 14,384 (75.7\%) & 5.26 & 17.87 & 5.53 & 17.38 \\
        Discard samples with low agreement & 97\%  & 9,561 (50.3\%) & 6.63 & 20.66 & 7.13 & 21.13 \\
        Discard samples with low agreement & 99\%  & 5,578 (29.3\%) & 7.98 & 22.76 & 8.51 & 23.48 \\
    \bottomrule
    \end{tabular}
\end{table*}
The results indicate that filtering out low agreement samples degrades performance. The CER increases from 4.95\% to 5.53\% when filtering out samples with less than 90\% agreement, reaches 6.63\% for a 97\% threshold, and 7\% for a 99\% threshold.
%At first sight, these results are surprising, as this approach ensures more reliable labels while maintaining a reasonable training dataset size.  A possible explanation is that removing low-agreement samples may lead to the removal of all difficult samples. 
Although this approach ensures more reliable labels, it also reduces the size of the training dataset, which usually leads to poorer performance. Another possible explanation is that removing low-agreement samples may lead to the removal of the most challenging samples. 

It should also be noted that PyLaia automatically discards invalid lines for CTC computation, which includes images with abnormally long transcriptions. As a result, a few unreliable samples are discarded before any agreement-based filtering.

\subsection{Impact of splitting strategy}
\label{sec:bias}
In this section, we discuss the limitations of our splitting strategy presented in Section 
\ref{sec:splitting_strategy}.

\subsubsection{Agreement based splitting}
We identify a bias introduced by our splitting strategy. The test set consists of lines with perfect agreement between annotators, which favors lines with little text and lines with legible handwriting. The same observation applies to a lesser extent to the validation set, which consists of lines with good agreement between annotators. 

\subsubsection{Random splitting}
In order to measure the impact of this bias, we train and evaluate another model on a random split, using only human annotations to allow comparison between the different sets. The random split is designed to have the same number of samples as the agreement-based split to measure the difference in terms of CER/WER. This corresponds to 19,013 images in the training set, 2,262 in the validation set, and 2,830 in the test set.

\todo{HM: Si on fait du random split, puis la moyenne, ça s'appelle de la cross-validation. Ca me semble le meilleur moyen de vérifier la stabilité des résultats avec des données incertaines.} 
\todo[color=yellow]{Solène: oui je suis d'accord, mais on n'a pas le temps de ré-entraîner plusieurs modèles (un entraînement = 24h)}

\subsubsection{Results}
Results of our experiments are presented in Table \ref{tab:results_split}.

\begin{table*}[tb]
    \centering
    \caption{Evaluation results using different splitting strategies}
    \label{tab:results_split}
    \begin{tabular}{l|ccccccc}
    \toprule
        \multirow{2}{*}{\bf Spliting strategy} & \multicolumn{2}{c}{\bf Train} &\multicolumn{2}{c}{\bf Validation} & \multicolumn{2}{c}{\bf Test} \\
        & CER (\%) & WER (\%) & CER (\%) & WER (\%) & CER (\%) & WER (\%) \\
    \midrule
        Agreement-based split & 8.74 & 23.85 & \textbf{6.05} & \textbf{22.18} & \textbf{5.57} & \textbf{19.12} \\
        Random split & \textbf{5.89} & \textbf{16.98} & 10.87 & 27.84 & 10.54 & 28.11\\
    \bottomrule
    \end{tabular}
\end{table*}

Using the agreement-based split, the test labels can be considered as gold transcriptions, as both annotators agree, which allows a fair assessment of performance. However, the results on this split clearly show a bias, as the performance on the training set is lower than on the test set. This suggests that test samples are "easier" than training samples. As a result, the model's performance on the test set is over-estimated.

On the random split, the CER and WER are much higher than on the agreement-based split. However, the test set may contain uncertain and noisy transcriptions, making it difficult to fairly assess the performance of the model. 

The actual performance of the model cannot be precisely assessed, it would be necessary to do cross-validation to estimate more precisely the performance of the model trained on the random split.
%but might fall between 5.57\% CER and 10.54\% CER.

\section{Conclusion}
\label{sec:conclusion}
In this paper, we have investigated different ways of training HTR models from crowdsourced annotations that may contain noise. We explored different training strategies to benefit from multiple uncertain annotations. The experimental results indicate that training a model with multiple noisy transcriptions or computing a consensus are two viable approaches. However, we observe that filtering training samples based on agreement between annotators introduces a bias in the training dataset and does not improve model performance. This observation suggests that data quantity may be more important than data quality, but it would be interesting to explore other quality metrics that do not introduce such a bias.

This project provides insights for improving annotation interfaces and for conducting future collaborative annotation campaigns. 
First, computing annotator-based confidence scores would allow campaign administrators to provide feedback to underperforming annotators. This information could also be used to train HTR models. 
Secondly, we believe that collaboration between annotators should be encouraged. For example, the integration of features that allow users to agree, vote, validate or correct existing transcriptions would be very useful.
Thirdly, the use of image quality metrics would help to remove blurred images and poorly segmented lines from the annotation campaign, as these tasks confused most annotators. 
Finally, the guidelines were not always followed by the annotators, so the addition of automatic validation checks would be useful. For example, suspicious words could be underlined, a warning could be issued if the text length does not match the line polygon area, or a Named Entity Recognition (NER) model could be used to flag non-capitalised names.
% an annotation tutorial and

For future work, we plan to explore Large Language Models (LLMs) to compute a clean consensus with corrected typing, grammar, and spelling errors. We would also like to explore training strategies for combining multiple transcriptions, such as using a custom annotator-weighted loss function or mechanisms similar to label smoothing.

%The results show that training a model directly using multiple noisy transcriptions is a viable solution. 
%\todo{HM: la quantité est plus important que la qualité?}
%However, selecting training samples based on the degree of agreement between annotators introduces a bias in the training data and does not improve the results. Data quantity seems more important than data quality. Could be interesting to explore other metrics for quality analysis that does not introduce such a bias.

%Other possible experiments: 1) try training strategies within models (custom loss function or something like label smoothing), 2) compute consensus with language model - ask to correct any typo. 

%Improving data quality - remove blurry images and badly segmented lines before asking people to annotate. 

%Reputation scores can also be computed for each annotator. This would allow us to 1) provide feedback for poorly-performing annotators, and 2) weight the different transcriptions using this score when training a model. 

%Improve the annotation interface 1) Encourage collaboration between annotators, let them agree/vote/correct the actual transcription 2) Add a validation/moderation phase. 3) Guidelines were not always followed by annotators. We could force people to run a tutorial or add automatic validation (ex: use a NER model to flag if names are not capitalized, underline words with a typo, add a warning is the text length is not consistent with the line polygon area)

\begin{acks}
We would like to thank the Municipal Archives of Belfort for giving us access to these documents, as well as the 87 annotators who made this project possible.
\end{acks}

\bibliographystyle{ACM-Reference-Format}
\bibliography{main}

%\appendix

\end{document}